\documentclass[a4paper, 10pt, conference, twocolumn]{article}                     

\usepackage{graphicx}
\usepackage{amsmath}
\usepackage{amssymb}
\usepackage{times}
\usepackage{fancyhdr}
\usepackage{csvsimple}
\usepackage[usenames, dvipsnames]{xcolor}
\usepackage[T1]{fontenc}
\usepackage[utf8]{inputenc}
\usepackage{hyperref}

\usepackage[backend=biber,natbib=true]{biblatex}

\date{}

\title{
	Learning to Label Affordances from Simulated and Real Data
}

\author{Timo Lüddecke and Florentin Wörgötter\\ \small Computational Neuroscience Group, Faculty of Physics, University of Goettingen}

\bibliography{literature}

\def\ADEe{\emph{ADE$^E$}}
\def\ADEt{\emph{ADE$^T$}}
\def\Sime{\emph{Sim$^E$}}
\def\Simt{\emph{Sim$^T$}}

\def\mathbbm#1{ \text{\textsc{1}} }

\def\noparagraph#1{\textit{#1}}

\begin{document}
	
\maketitle

\begin{abstract}
An autonomous robot should be able to evaluate the affordances that are offered by a given situation. Here we address this problem by designing a system that can densely predict affordances given only a single 2D RGB image. This is achieved with a convolutional neural network (ResNet), which we combine with refinement modules recently proposed for addressing semantic image segmentation. We define a novel cost function, which is able to handle (potentially multiple) affordances of objects and their parts in a pixel-wise manner even in the case of incomplete data. We perform qualitative as well as quantitative evaluations with simulated and real data assessing 15 different affordances. In general, we find that affordances, which are well-enough represented in the training data, are correctly recognized with a substantial fraction of correctly assigned pixels. Furthermore, we show that our model outperforms several baselines. Hence, this method can give clear action guidelines for a robot.
\end{abstract}

\begin{figure*}[h]
	\vspace{1cm}
	\centering
	\includegraphics[width=0.95\textwidth]{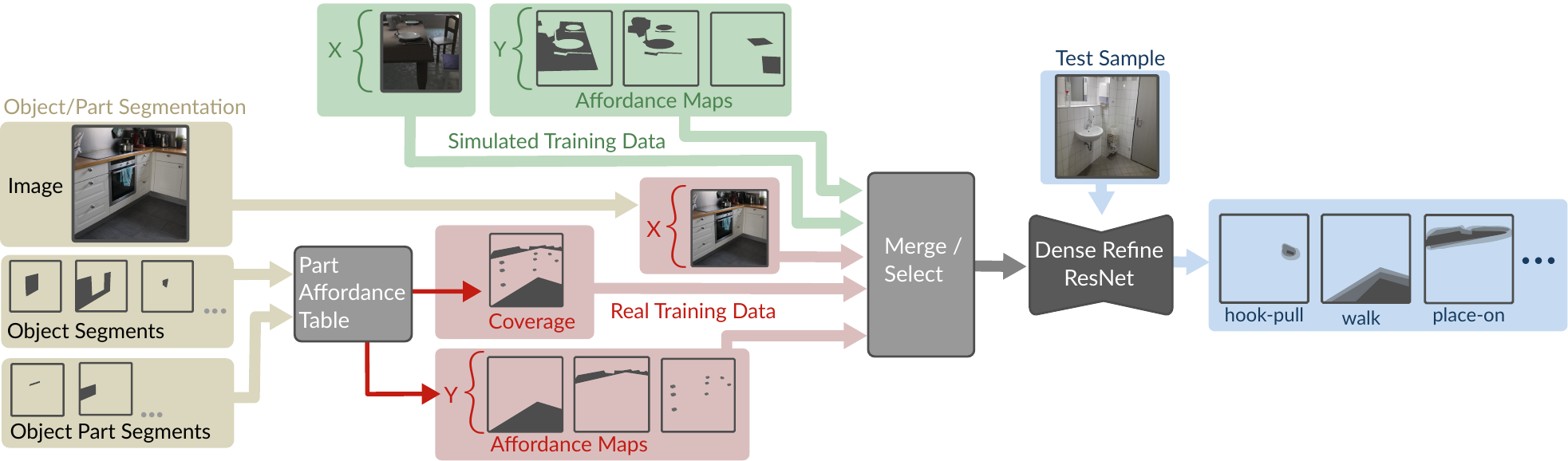}
	\caption{\label{method}Our approach: We train a neural network to predict a set of affordance maps (Y) from a single RGB image (X) using a new loss function that allows for incomplete data by incorporating a coverage map.}
\end{figure*}

\section{Introduction}

The goal of this study is to provide a deep learning-based system that allows a robot to distinguish different affordances in different visual scenes.

The term affordances originates from cognitive psychology \cite{gibson79} and means possible interactions between an animal and its environment. Examples for affordances from the perspective of a cat are: A bush affords shelter, a mouse affords nutrition. Later, the term was adopted by the robotics community and extended to from animals robots. Essentially in robotics this term very often takes the meaning of: ``Which actions could a robot perform in a given situation (with some given objects)?''

In this work, we assess affordances from the perspective of a human, which implies that such affordance may also be useful for a humanoid robot. In  addition, even human affordances, which a robot cannot directly make use of, may nonetheless help the machine to understand human behavior.

We focus on a set of $15$ affordances, which are listed in Table \ref{aff_set}. The goal is to predict a dense visual map for each of these affordances labeling the pixels in the scene with $1$ if a certain affordance is present and $0$ otherwise. These affordance maps then indicate, which actions could be performed with the different objects and places in a scene.

Central to our approach is that we perform affordance prediction by just using single RGB images. This 2D-approach allows using un-calibrated images of any kind and, thus, makes the method applicable in a wide range of robotics applications and other scenarios.

\section{Related Work}

\noparagraph{Dense Labeling}
From a methodological point of view, work in dense labeling is particular relevant, especially approaches that are based on neural networks.
As one of the first approaches, \citet{long15}, enhance and retrain the VGG16 network \cite{simonyan14} to densely predict class labels, with averaging up-sampled ``skip'' branches being a core idea.
The DeepLab model of \citet{chen15} adds a conditional random field to improve the alignment between predictions and edges in the image.
\citet{eigen15a} jointly predict depth, normals and object class labels by alternating between convolutions and incorporating multi-scale skip connections.
Segnet \citet{badrinarayanan17} store the pooling indices of the encoder and pass them to the up-sampling layers to preserve spatial accuracy.

Most relevant for our model is the work of \citet{pinheiro16}, which introduces refinement modules that integrate high- and low-level activations. The whole architecture is based on ResNet \cite{he16}.

\noparagraph{Affordances}
Affordances have been addressed in several works in the past (see for example Table~\ref{aff_comp}), both in robotics as well as in computer vision. A straightforward setting involves estimating affordances of whole objects, which is applied in the work of \cite{stark08} and \cite{zhu14}. Akin to these is the work of \citet{ye16} who detect bounding boxes of affordances using a two-stage approach consisting of region proposal and CNN-feature-based affordance recognition.

Affordances can also be predicted in form of (human) poses. This scheme is adopted by \cite{gupta11} making use of scene geometry and in \citet{grabner11} specifically for chairs and by \citet{fouhey14a} using video.

The idea of ``action maps'' is similar to affordance segmentation. However, the former
tends to be more specific, e.g. by referring to concrete objects and the set of considered actions is fairly small. Examples of these approaches are
\citet{savva14} who generate seven different ``action maps'' by tracking people in RGB-D video footage and \citet{rhinehart16} who learn 6 action maps through analyzing egocentric videos recordings.

The method proposed by \citet{roy16} is very similar to ours as it also generates pixel-wise maps given an RGB image. Their model learns intermediate representations for depth, surface normals and object classes which are then employed to carry out the affordance map prediction. The learning of these representations is actively enforced during training, i.e. the method requires additional data during training, while our method only needs RGB images and affordance map ground truth. Another difference to our work is the set of considered affordances.

\begin{table}[t]
	\footnotesize
	\vspace{0.3cm}
	\centering
	\begin{tabular}{llll}
		\hline
		Approach & \# & input & output \\
		\hline
		\citet{grabner11}  & 1 & RGB-D & per voxel \\
		\citet{gupta11} & 4 & RGB & per pixel \\
		\citet{savva14} & 7 & Video & per voxel \\
		\citet{rhinehart16} & 6 & Video & per grid-cell \\
		\citet{roy16} & 5 & RGB & per pixel \\
		\hline
		our approach & 15 & RGB & per pixel \\
		\hline
	\end{tabular}
	\vspace{0.2cm}
	\caption{\label{aff_comp}Comparison of related algorithms with \# denoting the number of used affordances.}
	\hfill
\end{table}

\section{Methods}

\subsection{Part Labels for Affordance Definition}
\label{from_object} 
We use real as well as simulated data for training (and testing) our system. In the following we will describe how to assign affordances to real scenes, which is the more complicated case. Generation of simulated scenes is described afterwards, where the same principles for affordance assignment are employed.

Our method relies on human-provided labels for the affordances of object parts for training. We use 15 types of affordances (Tab.~\ref{aff_set}) and define them using three guiding principles:

\begin{table}[t]
	\footnotesize
	\centering
	\setlength{\tabcolsep}{0.2em}
	\renewcommand{\arraystretch}{1.2}
	\begin{tabular}{l|lllllllllllll}
		object & \rotatebox{90}{obstruct} & \rotatebox{90}{pinch-pull} & \rotatebox{90}{break} & \rotatebox{90}{sit} & \rotatebox{90}{grasp} & \rotatebox{90}{illumination} & \rotatebox{90}{support} & \rotatebox{90}{place-on} & ...  \\
		\hline
		*/knob     & 1 & 1 & 0.5 & 0   & 1   & 0 & 0 & 0 & ...\\
		*/top      & 0 & 0 & 0 & 0.5   & 0   & 0 & 0 & 1 & ...\\
		pot        & 1 & 0  & 0.5 & 0   & 1   & 0 & 0 & 0 & ...\\
		\rotatebox{90}{...} & \rotatebox{90}{...} & \rotatebox{90}{...} & \rotatebox{90}{...} & \rotatebox{90}{...} & \rotatebox{90}{...} & \rotatebox{90}{...} & \rotatebox{90}{...} & \rotatebox{90}{...} & \rotatebox{135}{...} \\
	\end{tabular}
	\vspace{0.2cm}
	\caption{
		\label{transfer}
		Excerpt from the transfer table. (0/0.5/1) = affordance (not/partial/fully) present, * = joker, that matches any object.}
\end{table}

\noindent
\begin{enumerate}
    \item 
Affordances should be \emph{valuable} (in some sense) for robots or humans.

\item
We require that affordance names are specific. For example, \emph{open} is a very unspecific multi-action. A \emph{pinch-pull}, on the other hand, can be used for opening a container (and also for doing other things). Hence, if an action is considered then we will try to use the corresponding most specific action descriptor (action word) to annotate the affordance.

\item
Actions can have a hierarchy, but lead to the same final outcome: E.g. a house can be entered, a door, which is a part of the house, can be opened and the door's handle, as a part of the door can be pulled. All of this will be done to enter the house, where the pulling of the door handle is here the action at the lowest semantic hierarchical level. Only this level will be considered to label affordances in this study. The other levels could possibly be addressed from there on using a reasoning- or inference-engine, which is not part of this work.
\end{enumerate}

Adhering to these principles we create affordances for network training using the following methods.

\noparagraph{Object Parts}
An affordance most often refers to only a part of an object. For example, it is the surface of the table that affords placing an object there, but not the table legs. Thus, we define affordances part-wise. The novel ADE20K dataset \cite{ade_20k} is currently the only one that resolves objects into their parts. Hence ADE20K will be used here.

\noparagraph{Transfer Table}
We use a manually defined look-up table for the mapping from object labels to affordance maps. The transfer table (Tab.~\ref{transfer}) uses the following annotation hierarchy:
object names (like ``cabinet''), paths to object-parts (like ``cabinet/drawer/handle'') or pure parts (like ``*/drawer''). These annotations are associated with different 15-dimensional affordance vectors, where every dimension corresponds to one affordance. Clearly, multiple affordances can be present simultaneously.

This annotation hierarchy is searched from specific to general to associate a concrete object or part to an affordance. Hence, in the example above, we first ask if ``cabinet/drawer'' is specified in the table and only if it this not found ``*/drawer'' is checked.

From the ADE20K dataset, we only use the most frequent 500 objects and parts.  For very large objects multiple affordances can exist; most, if not all, only apply to specific parts (and not to the whole object). In addition, for some parts of the image, no affordance may exist. This case is treated in the cost function (see section \ref{cost_function}), exploiting the fact that we know \emph{where} affordance-data is missing.

\begin{table}
	\setlength{\tabcolsep}{0.3em}
	\footnotesize
	\renewcommand{\arraystretch}{1.2}
	\begin{tabular}{|l|p{5.8cm}|}
		\hline
		\textbf{Affordance} & \textbf{Description}                                                                                                       \\ \hline
		obstruct           & vertical surface that prevents locomotion. \textit{e.g. wall}                                                              \\
		break              & detachable objects that can easily be damaged or destroyed \textit{e.g. vase }   \\
		sit                & surface a human can sit on while having the feet on the ground \textit{e.g.  seat cushion}                                                                    \\
		grasp              & detachable objects that can be encompassed with one hand or only few fingers and be moved with one arm.\textit{e.g.  vase)} \\
		pinch-pull         & surfaces that can be pulled through a pinch movement (all directions). \textit{e.g.  knob}                                                   \\
		hook-pull          & surfaces that can be pulled by hooking up fingers (all directions). \textit{e.g.  handle }  \\
		tip-push           & surfaces that trigger some action when being pushed. \textit{e.g.  button-panel }                                                  \\		
		warmth             & surfaces that emit warmth. \textit{e.g. fireplace  }                                                                        \\
		illumination       & surfaces that emit visible light.\textit{e.g. bulb }                                                                      \\
		observe            & surfaces that present information or art, i.e. that can be read or watched. \textit{e.g.  display}                                                           \\
		support            & stable surfaces that provide support for standing (for the agent) except ground. \textit{e.g. wall}\\
		place-on           & raised surfaces where objects can be placed on (this excludes the ground). \textit{e.g.  tabletop}                                                             \\
		dry                & surfaces capable of soaking up water. \textit{e.g.  towel}                                                                      \\
		roll               & surfaces that can be used with wheels. \textit{e.g.  road}                                                                 \\
		walk               & surfaces a human can walk on. \textit{e.g. grass} \\
		\hline
	\end{tabular}
	\caption{\label{aff_set}Description of the used set of affordances.}
\end{table}

\noparagraph{Data Augmentation}
Scene quality in ADE20K substantially varies. This leads to a situation that only a rather small number of good-quality training samples can directly be generated from ADE20K. Therefore, we augment the dataset by cropping  out image patches from an original image where we then vary color and contrast within such a patch. How many crop one can obtains depends on the original image quality. Better images with many objects can be used to create more augmented data.

In remainder of this paper, the training-split of this dataset, involving 17955 samples, is referred to as \ADEt and the evaluation-split which contains 1970 samples is named \ADEe.

\subsection{Simulation Model}
Transferring labels from real-image object parts has some disadvantages: Maps are incomplete and some affordances occur rarely. We overcome this problem by generating a new dataset of simulated images. It relies on a probabilistic scene model of a living room and a kitchen with several constituents of the scene being randomized. Hence, we can generate strongly varying images of the scene. More precisely, the randomized variables in our model are object material, -position, -shape, scene illumination, and perspective.

\noparagraph{Object material}
Objects can have different materials. A table surface, for instance, can be composed of plastic, wood or glass. Glass can be transparent or opaque. During scene generation, every object in the scene gets a randomly assigned material, with possibilities being constrained based on the object name.

\noparagraph{Object positions}
Several objects are randomly positioned in the scene and relative to other objects. Examples are a plate on a table, which is dependent on the table's height and a fork and knife, which are positioned relative to the plate.

\noparagraph{Object shape}
For some objects we define key model shapes and interpolate between these key shapes when a scene is generated. E.g. we interpolate between a chair with rounded edges and a chair with sharp edges.

\noparagraph{Scene illumination}
The world during day-time looks entirely different than at night. We account for this by varying light from the outside as well the intensity of indoor and outdoor illumination.

\noparagraph{Perspective}
Having obtained a variable scene model, we still need to simulate the process of photography by projecting the 3D scene onto a 2D plane from many possible viewpoints. For this, it is desirable to use viewpoints that sample mostly interesting aspects of the scene (e.g. multiple objects and sufficient distance), while avoiding irrelevant projections (e.g. view of the ground only) or invalid perspectives (e.g. taking an image from behind a wall).
We address this challenge by sampling the camera's position randomly along a fixed heuristically assumed trajectory and introducing slight variances with respect to the position.

For each object or object part we manually define corresponding affordances and render the corresponding affordance maps in a second pass by changing the objects' materials.

This procedure allows us to generate an arbitrary number of training samples each providing consistent, fully covered affordance maps. This way, we can extend the training set by many additional images. 

\begin{figure}
	\vspace{0.25cm}
    \centering
	\includegraphics[width=1.4cm]{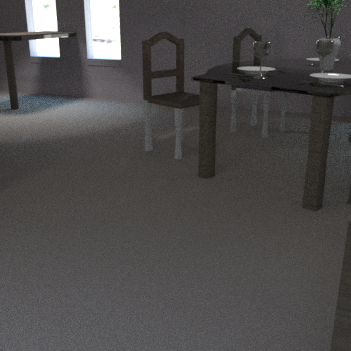}
	\includegraphics[width=1.4cm]{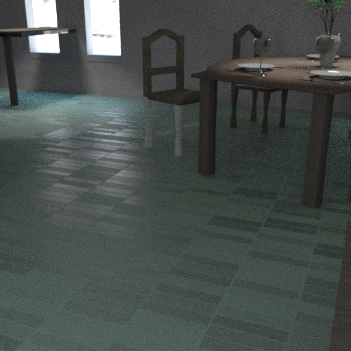}
	\includegraphics[width=1.4cm]{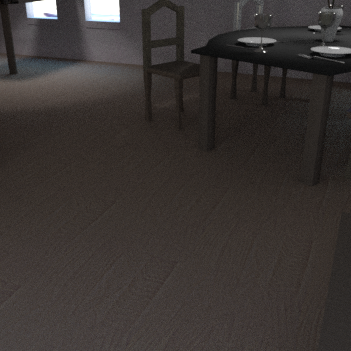}
	\includegraphics[width=1.4cm]{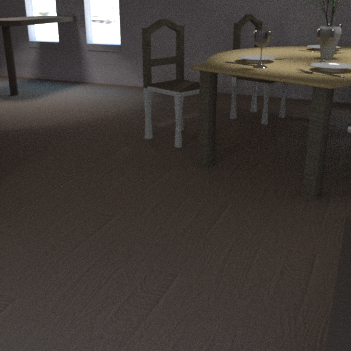}	
	\includegraphics[width=1.4cm]{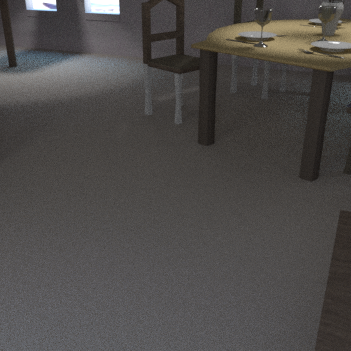}
	\caption{\label{sim_samples}Multiple images generated from a fixed perspective while other variables were randomly sampled.}
\end{figure}

The simulation model is implemented in the open source 3D modeling and simulation software blender\footnote{https://blender.org} using it's scripting API and the unbiased, physics-based renderer cycles. Figure \ref{sim_samples} gives an impression of the variability of the simulated samples. Scenes obtained using this method are subsequently denoted by \Simt.

\subsection{CNN}

\begin{figure}
	\centering
	\vspace{0.3cm}
	\includegraphics[width=5cm]{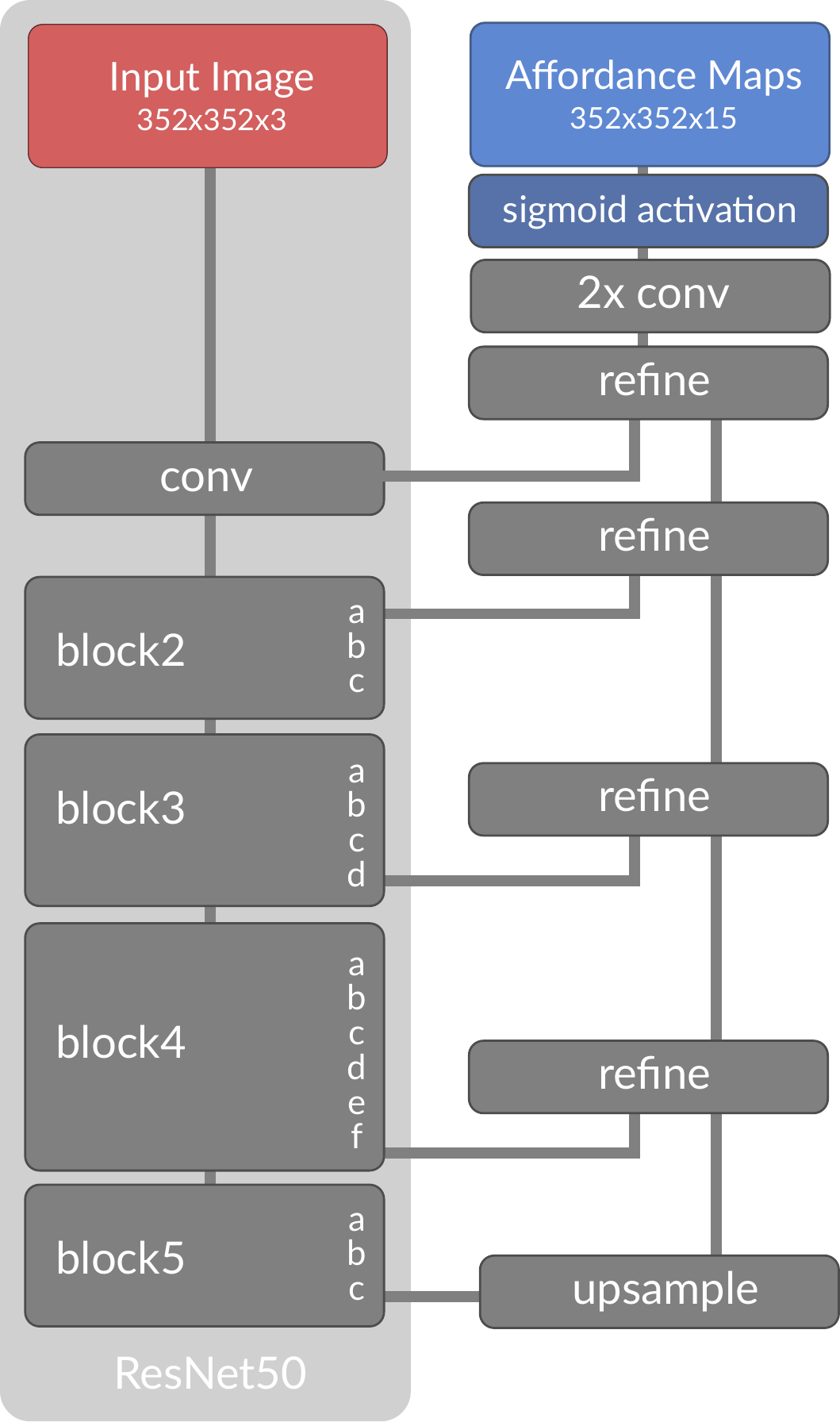}
	\caption{\label{model}Architecture of our proposed Dense-Refine-ResNet. With the lowercase letters indicating sub-blocks within the ResNet50 architecture. }	
\end{figure}

\noparagraph{General Definition}
Affordances are context dependent. An example makes this clear.  We could ask whether a surface is walk-able or suitable to place things? If we now compare the ground with a table surface, we find that, locally, both are flat and uniform. Only context may resolve the difference between them. Walk-able surfaces, for example, may be accompanied by cars and trees, a table surface, on which we would put things --- on the other hand --- is often flanked by e.g. chairs. This leads to the requirement that the receptive field of a pixel should, ideally, cover the whole image because even distant pixels might be decisive for a local affordance.

This could well be in conflict with the second essential requirement, which demands that image details must not get lost during the forward pass of the network. Hence, object- and part-boundaries should be preserved. For example, many affordances concern rather smaller image aspects (e.g. a knob for pulling) and these aspect should not be lost by the network's operation. With these requirements in mind, we propose a novel neural network architecture, depicted in Figure \ref{model},  that is based on ResNet50 \cite{he16} and adopts the idea of refinement modules \cite{pinheiro16}. It had been shown by these authors that ResNet50 together with refinement modules successfully generates object proposals, because refinement modules offer an elegant way for merging local with scene-level information. Thus, here we use a modified version of the architecture in \cite{pinheiro16}.

This architecture then integrates abstract information from deep layers with the spatially more accurate representations still present in less deep layers. Here both input layers will deliver maps of the same image size where they are then first stacked on top of each other (concatenated along depth) and subsequently convolved with the learned filters to obtain $15 k$ feature maps. The variable $k$ is a control parameter of our model (see Section~\ref{experiments}). To reduce training effort, we preserve the weights learned from ImageNet in the original ResNet encoder and only train the refinement modules (see ``encoder train'' parameter in Table \ref{quant}).

\noparagraph{Cost function}
\label{cost_function}
Here we propose a novel cost function we call masked binary cross entropy. This cost functions deals with two aspects: 1) affordances are often not unique and for a given pixel multiple affordances may exist. Hence, we imply a binary (present vs not present) probability distribution for each pixel and each affordance. 2) Some parts of the image may contain no (or indecisive) affordances. For those, we can not tell whether an affordance is present or not, because the corresponding object or part is not found in the transfer table. However, since we generated the affordance maps, we know the location of the invalid regions. The idea is to incorporate also this information into the cost function.

Both aspects from above lead to the fact that commonly-used cost functions for semantic segmentation cannot be employed here.

Subsequently, we will formally derive the here used \emph{masked binary cross entropy cost function}.

\def\a{\mathbf{Y}}
\def\b{\mathbf{\hat{Y}}}
\def\sumimg{\sum_{i \in \mathcal{I}}}
\def\sumaff{\sum_{a \in \mathcal{A}}}
\def\ent{H(\a_{ai}, \b_{ai})}
\def\mask{\mathbf{M}_{i}}

\label{masked_loss}
We annotate the ground truth matrix of an image for affordance $a \in \mathcal{A}$ and pixel $i \in \mathcal{I}$ with $\a_{ai}$ and the associated model prediction is given by $\b_{ai}$.

Then the binary cross entropy $H$ is defined by:
$ H(p,q) = - p \log{(q)}\ - \ (1\!-\!p)\log{(1\!-\!q)}$.
This is summed up to render a scalar loss (cost), which captures the average binary entropy over all affordances and the image.
\[
\mathcal{L}(\mathbf{\a,\b}) = (|\mathcal{A}||\mathcal{I}|)^{-1} \sumaff \sumimg \ent
\]
So far this definition is compatible with non-exclusive classes, but it does not yet account for incomplete data. To achieve this, we mask the cross entropy matrix, excluding all regions where no (or indecisive) affordance is present, before averaging. Masking is a very efficient and simple way for removing the incompleteness ambiguities and we get the following loss:
\[\mathcal{L}^{\text{m}}(\mathbf{\a,\b}) = (|\mathcal{A}| \sumimg \mask)^{-1} \sumaff \sumimg \mask \ent \]
with $\mask \in \{0,1\}$ indicating if pixel $i$ is valid, i.e. if a corresponding entry is found in the transfer table.

\begin{table*}[h!]
		\vspace{0.25cm}
	\centering
	\footnotesize
	\setlength{\tabcolsep}{0.2em}
	\renewcommand{\arraystretch}{1.2}
	\begin{tabular}{|lcccccc|cc|cc|cc|}
		\hline 
		& &  \bfseries encoder & \bfseries masked & & & \bfseries training & \multicolumn{2}{c|}{\bfseries mean IoU } & \multicolumn{2}{c|}{\bfseries mean accuracy } & \multicolumn{2}{c|}{\bfseries pixel accuracy } \\
		\bfseries Network  &
		\bfseries k &		
		\bfseries train &
		\bfseries loss &
		\bfseries epoch &
		\bfseries val. error &
		\bfseries dataset &
		\bfseries \Sime & \bfseries \ADEe
		& \bfseries \Sime & \bfseries \ADEe
		& \bfseries \Sime & \bfseries \ADEe \\
		\hline\hline	
		ResNet50+Refine (R1) & 5 & - & - & 21 & 0.119 & \ADEt & \color{Maroon}0.222 & 0.284 & \color{Maroon}\bfseries 0.529 & 0.559 & \color{Maroon}0.782 & 0.803 \\ 
ResNet50+Refine (R2) & 5 & \checkmark & \checkmark & 14 & 0.130 & \ADEt & \color{Maroon}0.221 & \bfseries 0.285 & \color{Maroon}0.502 & 0.605 & \color{Maroon}0.781 & 0.865 \\ 
ResNet50+Refine (R3) & 3 & \checkmark & \checkmark & 11 & 0.128 & \ADEt & \color{Maroon}\bfseries 0.227 & 0.283 & \color{Maroon}0.509 & 0.607 & \color{Maroon}0.757 & 0.855 \\ 
ResNet50+Refine (R4) & 5 & \checkmark & \checkmark & 14 & 0.130 & \ADEt & \color{Maroon}0.219 & 0.284 & \color{Maroon}0.505 & \bfseries 0.626 & \color{Maroon}\bfseries 0.792 & \bfseries 0.870 \\   \hline  \hline
ResNet50+Refine (R5) & 3 & \checkmark & - & 09 & 0.056 & \Simt & 0.611 & \color{MidnightBlue}0.101 & 0.695 & \color{MidnightBlue}0.168 & 0.934 & \color{MidnightBlue}0.499 \\ 
ResNet50+Refine (R6) & 5 & \checkmark & - & 11 & 0.060 & \Simt & 0.660 & \color{MidnightBlue}0.108 & 0.717 & \color{MidnightBlue}0.170 & 0.930 & \color{MidnightBlue}0.516 \\ 
ResNet50+Refine (R7) & 3 & - & \checkmark & 28 & 0.024 & \Simt & \bfseries 0.833 & \color{MidnightBlue}\bfseries 0.126 & \bfseries 0.883 & \color{MidnightBlue}\bfseries 0.198 & \bfseries 0.976 & \color{MidnightBlue}\bfseries 0.518 \\   \hline  \hline
ResNet50+Refine (R8) & 3 & \checkmark & \checkmark & 12 & 0.117 & \Simt + \ADEt & 0.487 & 0.275 & 0.674 & 0.435 & 0.900 & 0.700 \\
ResNet50+Refine (R9) & 3 & - & \checkmark & 03 & 0.112 & \Simt + \ADEt & 0.505 & \bfseries 0.295 & 0.672 & 0.462 & 0.896 & 0.723 \\ 
ResNet50+Refine (R10) & 3 & - & - & 21 & 0.107 & \Simt + \ADEt & \bfseries 0.597 & 0.247 & \bfseries 0.736 & 0.362 & \bfseries 0.932 & 0.581 \\ 
ResNet50+Refine (R11) & 5 & - & \checkmark & 01 & 0.122 & \Simt + \ADEt & 0.463 & 0.262 & 0.650 & \bfseries 0.464 & 0.887 & 0.733 \\  \hline
VGG16+Upsampling & 3 & - & \checkmark & 20 & 0.135 & \Simt + \ADEt & 0.399 & 0.277 & 0.594 & 0.444 & 0.834 & \bfseries 0.749 \\ 
VGG16+Upsampling & 5 & - & - & 16 & 0.129 & \Simt + \ADEt & 0.443 & 0.238 & 0.654 & 0.380 & 0.850 & 0.593 \\ 
VGG16+Upsampling & 3 & - & - & 28 & 0.130 & \Simt + \ADEt & 0.423 & 0.255 & 0.592 & 0.374 & 0.846 & 0.634 \\ 
VGG16+Upsampling & 5 & - & \checkmark & 14 & 0.130 & \Simt + \ADEt & 0.434 & 0.282 & 0.625 & 0.450 & 0.852 & 0.734 \\  \hline
\multicolumn{5}{|l}{SegNet} & 0.120 & \Simt + \ADEt & 0.518 & 0.290 & 0.711 & 0.470 & 0.914 & 0.721\\
		\hline			
	\end{tabular}
	\caption{\label{quant}Quantitative Results of our method in comparison with state-of-the-art algorithms.}
\end{table*}

\section{Experimental Setup}
\label{experiments}

\noparagraph{Evaluation Datasets}
The training and validation samples of \ADEt are generated from a 90\% portion of the ADE20K \emph{training} dataset. From the remaining 10\% we manually pick 50 images of good quality, transfer the annotations to affordances and let an expert manually correct this according to the definitions provided by Table \ref{aff_set}. Due to this manual correction, systematic errors of the part-to-affordance conversion procedure are punished during evaluation and we obtain a more realistic estimate of the error. Additionally, we test on simulated data which was generated the same way as the simulation training data.

Consistent to the training datasets, the evaluation datasets are denoted as \ADEe and \Sime.

\noparagraph{Metrics}
For quantification,  we use measures which are common in the field of semantic segmentation. We measure: \\
(1) pixel accuracy  {\small \begin{equation*} 
\text{pixel Acc}(\a, \b) = \frac{\sumaff \sumimg \mathbbm{1} [\a_{ai} = \b_{ai}] }{\sumaff \sumimg \mathbbm{1} [\a_{ai} = 1] }, \end{equation*} }
(2) mean class-wise accuracy  {\small\begin{equation*} \small \text{mean Acc}(\a, \b) =\frac{1}{|\mathcal{A}|} \sumaff \frac{\sumimg \mathbbm{1}[\a_{ai} = \b_{ai}]}{\sumimg \mathbbm{1} [\a_{ai} = 1]} \text{ and} \end{equation*} }
(3) mean class-wise intersection over union (IoU) {\small \begin{equation*}
 \text{mean IoU}(\a, \b) =\frac{1}{|\mathcal{A}|} \sumaff \frac{\sumimg \mathbbm{1}[\a_{ai} = 1 \land \b_{ai} = 1]}{\sumimg \mathbbm{1} [\a_{ai} = 1 \lor \b_{ai} = 1]},
\end{equation*}} following the notation introduced in section \ref{cost_function} with $\a$ denoting ground truth and $\b$ a model's prediction.
$\mathbbm{1}[]$ is the indicator function.

Of central relevance in this study will be the IoU measure also known as Jaccard Index. Maximal IoU would be 1.0. It is important to note that IoU is measuring the overlap with the ground-truth image segment area and is punished for both, lack of overlap in the labeling as well as false positive outside-of-segment labeling.

\noparagraph{Output Binarization}
To compute metrics, the probabilistic predictions of the network must be binarized. We determine optimal thresholds for each affordance by considering IoU scores on the \Simt dataset and simply selecting the threshold with the best performance.

\noparagraph{Implementation Details}
For training and evaluations we employ Geforce Titan X, Geforce 1080 Ti and Geforce 1060 GTX GPUs and \cite{keras} with \cite{tensorflow} backend. Gradients are updated using the RMSprop method \cite{rmsprop} and training is stopping if the validation error did not improved for 5 epochs.

\noparagraph{Baselines}
To judge the performance of our model we compare against two baseline models: VGG16 + Upsamping is a simple encoder-decoder (or hourglass) architecture without skip connections.

Additionally, we re-implement the well-known SegNet \cite{badrinarayanan17} architecture and adapt it to affordance segmentation by using sigmoid activations instead of softmax. In contrast to the simple encoder-decoder architecture this method memorizes the pooling indices and makes use of this information during the upscaling-phase.

\section{Results and Discussion}
\subsection{Quantitative Analysis}

Table \ref{quant} presents the results of our evaluation on different configurations of the proposed method with baseline models. The columns of this table report on detailed configurations of the respective model, involving parameter $k$, indication if the encoder weights were trained (encoder train), the used loss function (masked or normal binary cross entropy), the epoch at which the respective model was chosen and the corresponding error on the validation dataset. 

We experimented also with dropout, separately applied in the encoder and decoder as well as applied in both, but find that these variations do not lead to substantial improvements, while slowing down the training procedure. Therefore, dropout is not considered any further.

Note, we used the often applied approach of stopping in the training evaluating the validation error. If this does not improve for 5 epochs training is stopped because then the network has basically converged. This is the only fair point for model comparison.

We consider intersection-over-union (IoU) to be the most meaningful metric as it requires very good predictions (low false negative \textit{as well as} low false positives) to obtain high scores and consequently we focus on IoU during this discussion. Other metrics are reported in Table \ref{quant} for completeness.

Rows are grouped according to the training dataset used and best results are highlighted in bold.

\noparagraph{General Observations}
All values reported are statistically highly significant due to the large-enough database used. We took care to find for the baseline methods (VGG16, SegNet) the optimal performance parameters. Still, all baselines produces lower values than our approach. Furthermore, as expected, in all real data cases we get smaller IoUs , whereas for simulated data IoUs are larger. Cross-modality learning, shown by the red and blue entries, demonstrates the across-dataset generalization capabilities of the network, i.e. how well a network trained on one dataset performs on the other. Blue entries are for $\Simt ~\rightarrow~ \ADEe$ and red ones for $\ADEt ~\rightarrow~ \Sime$. Models trained using rendered images tend to perform badly on real data while the opposite direction works well. This is likely due to the variance being larger in real data and models that were trained on simulated data would have to extrapolate beyond their training distribution.

\noparagraph{Model Complexity Parameter $k$, Column 1}
Regarding model complexity, we do not see any difference between $k=3$ and $k=5$, where $k=1$ and $k=7$ in general produced worse results. This suggests that moderately small models have a sufficient number of parameters to fit the data well. 

\noparagraph{Encoder Training, Column 2}
Furthermore, we found that it is better to train the decoder only and keep the encoder's weights fixed. This can be seen in Table \ref{quant} for models R8 and R9.

All in all, training without any ``bells and whistles'' leads to the best performance.

\noparagraph{Masking, Column 3}
The masked cost function introduced in \ref{masked_loss} does not have a big impact on the quality of the predictions but one can see that the IoU scores on \ADEe when trained on  \Simt + \ADEt tend to be higher if masked binary cross entropy was applied. This makes intuitively sense as the simulated data is complete anyway and masking can only improve training on \ADEt.

\noparagraph{Joint Training, Rows R8-R11}
In addition to models trained on individual datasets, we also train models on both datasets conjointly: The datasets are first concatenated and then randomly mixed. This way, each mini-batch for training is composed of samples from both datasets and each update of the network weights through the gradient will reflect this. Hence, the network learns both datasets simultaneously.

In our experiment, we find that joint training improves performance. Scores on \ADEe are on-par with scores of models trained on \ADEt only while the performance on \Sime is, as expected, significantly better ($>0.5$ instead of $\approx 0.24$).

\begin{figure*}
		\vspace{0.25cm}
	\centering
	\includegraphics[width=2.4cm, height=2.4cm]{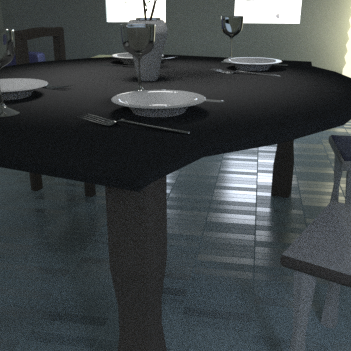}
	\includegraphics[width=2.4cm, height=2.4cm]{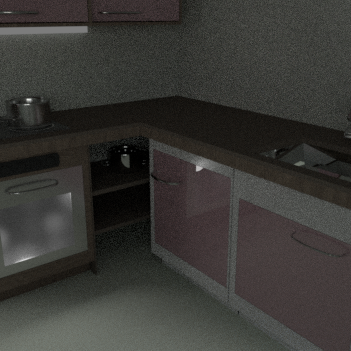}
	\includegraphics[width=2.4cm, height=2.4cm]{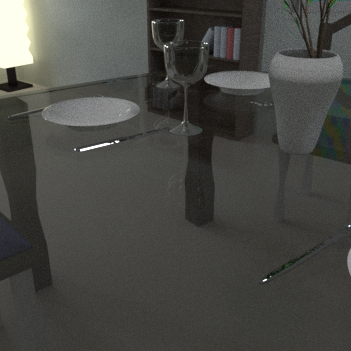}
	\includegraphics[width=2.4cm, height=2.4cm]{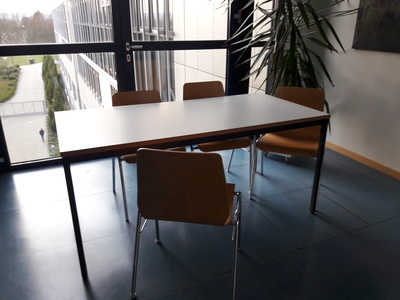}
	\includegraphics[width=2.4cm, height=2.4cm]{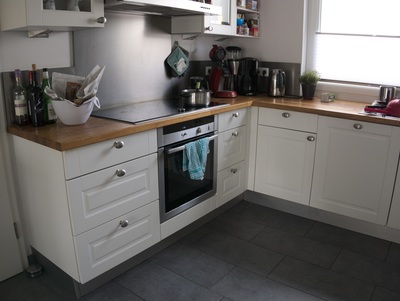}
	\includegraphics[width=2.4cm, height=2.4cm]{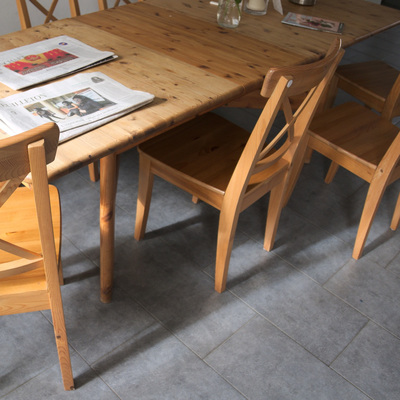} \\[0.1cm]	
	\includegraphics[width=2.4cm]{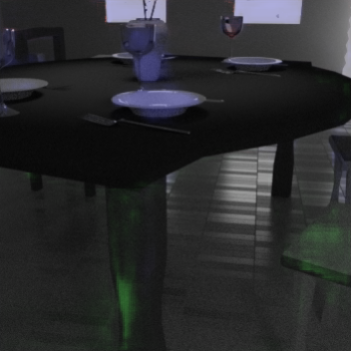}
	\includegraphics[width=2.4cm]{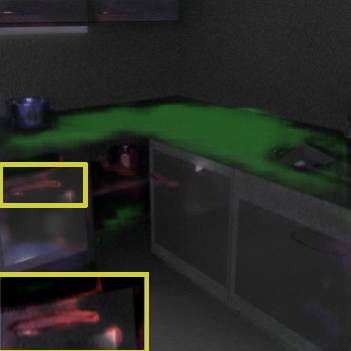}
	\includegraphics[width=2.4cm]{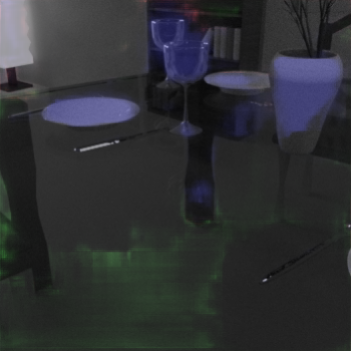}			
	\includegraphics[width=2.4cm]{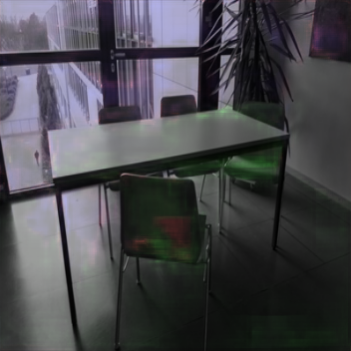}
	\includegraphics[width=2.4cm]{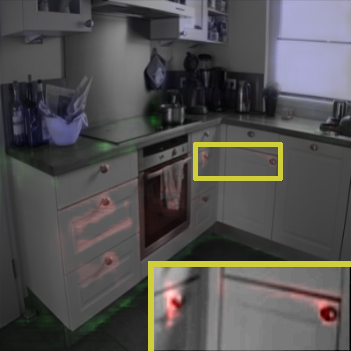}	
	\includegraphics[width=2.4cm]{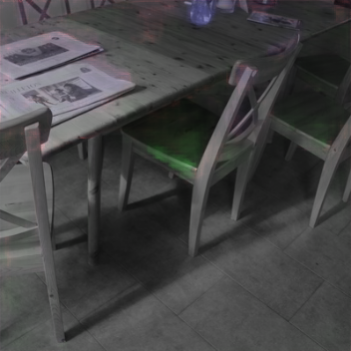} \\[0.1cm]
	\includegraphics[width=2.4cm]{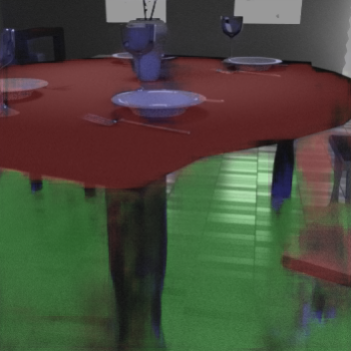}	
	\includegraphics[width=2.4cm]{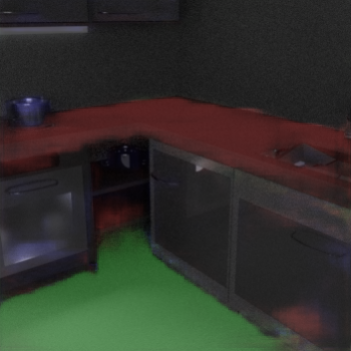}	
	\includegraphics[width=2.4cm]{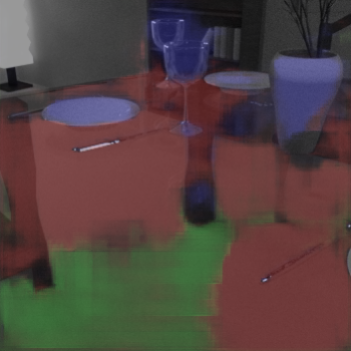}		
	\includegraphics[width=2.4cm]{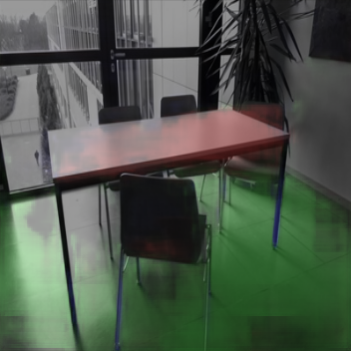}	
	\includegraphics[width=2.4cm]{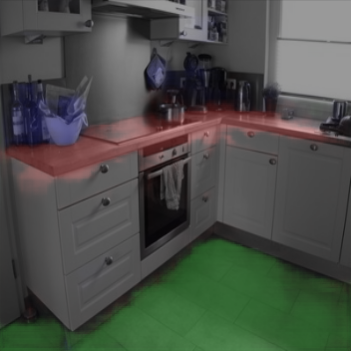}			\includegraphics[width=2.4cm]{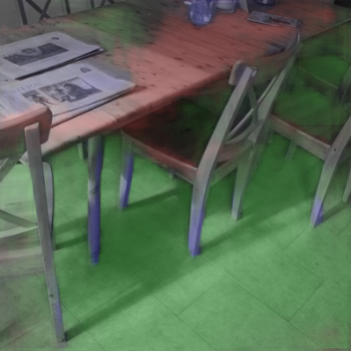}	
	\centering
	\caption{\label{quali}Qualitative assessment of model R9. The three columns on the left show simulated images, while the right columns use photographs. In some cases, details are magnified in insets with increased contrast.
		Top row: input image,
		middle: {\color{red} hook-pull}, {\color{OliveGreen} sit}, {\color{MidnightBlue} break},
		bottom: {\color{red} place-on}, {\color{OliveGreen} walk}, {\color{MidnightBlue} grasp}. Note, in case of simultaneous presence of two affordances the colors blend.
	}
\end{figure*}

\noparagraph{Affordance-wise Evaluation}
Table \ref{indiv} reports individual scores for the model R10 of Table \ref{quant}.

\begin{table}
	\centering
	\setlength{\tabcolsep}{0.2em}
	\footnotesize
	\renewcommand{\arraystretch}{1.2}
	\begin{minipage}[t][][c]{3.5cm}
	\begin{tabular}{|l|cc|}
		\hline
		& \multicolumn{2}{c|}{ \bfseries IoU} \\
		\bfseries affordance &  \bfseries \Sime & \bfseries \ADEe \\
		\hline
            obstruct & 0.929 & 0.826  \\
            break & 0.660 & 0.405 \\
            sit & 0.455 & 0.182 \\
            grasp & 0.563 & 0.153 \\
            pinch\_pull & 0.046 & 0.003 \\
            hook\_pull & 0.084 & 0.050 \\
            tip/push & 0.000 & 0.002 \\
            warmth & 0.220 & 0.009 \\
		\hline
	\end{tabular}
	\end{minipage}
	\begin{minipage}[t][][c]{3.5cm}
		\begin{tabular}{|l|cc|}
			\hline
			& \multicolumn{2}{c|}{ \bfseries IoU} \\
			\bfseries affordance &  \bfseries \Sime & \bfseries \ADEe \\
			\hline
            illumination & 0.609 & 0.455 \\
            read/watch & 0.828 & 0.147 \\
            support & 0.760 & 0.623 \\
            place\_on & 0.689 & 0.112 \\
            dry & 0.026 & 0.238 \\
            roll & 0.850 & 0.611 \\
            walk & 0.850 & 0.608 \\		
			\hline
		\end{tabular}
	\end{minipage}
	\caption{\label{indiv}Performance of individual affordances.}
\end{table}

This shows that affordances that often occur and cover large areas tend to be learned more reliably. This is also the reason for the poor performance of some rare, small affordances (pull). 

\subsection{Qualitative Evaluation}
In Figure \ref{quali} we show predictions of network R10 in Table \ref{quant} after training for 21 epochs on both datasets. For these results we do not use binarization. All colored pixels encode the probability for the corresponding affordance by color intensity. This renders an assessment of the degree of confidence the model attains for any given pixel's affordance. Mixed colors indicate the presence of mixed affordances. The here-used images challenge the network with difficult situations like front-lighting and transparent materials.  It is remarkable that, indeed, all affordance predictions appear to be reasonable. Furthermore, we find that the performance on novel images is good, although the model's confidence is in these cases lower.  In particular for small and rare structures like the knobs in the 5th column: There is only a very weak signal, which is barely recognizable in the image. This is likely due to the cost function weighting all (or all valid) pixels equally.

Summed up, the qualitative samples confirm the observations from the quantitative evaluation: The method works well as long as enough training data is available.


\section{Conclusion}
In this paper we have described a method that labels a comparatively large set of 15 affordances pixel-wise given only single RGB images. We show that state-of-the art semantic segmentation methods can be adapted to this new task with simple modifications only and learn to produce good affordance maps if enough data of good quality is provided.

In all cases trained on real images we found that at least a best-performing IoU of more than 0.2. Note that these numbers depend on the binarization threshold. For example, a lower threshold will increase the labeled pixel percentage but at the cost of increasing false positives, too. A higher threshold leads to less false positives but all to a lower correctly labeled set of pixels. Both would lead to a reduced IoU.

From a robotic perspective it might, however be advisable to increase the threshold (going away from an optimal IoU), because this way one obtains a labeling, which is sparse but highly reliable and avoids false positives. Why would this be better for robot action selection? Avoiding/reducing false positives reduces the danger of potentially damaging action-choices by a robot. Different from this, a low number of reliably labeled pixels will not negatively influence the actions of a machine. This is due to the fact that semantic 3D part segmentation algorithms exist that extract parts with a single action relevant meaning (like "handle", "blade", etc., \cite{schoeler16}). The here presented affordance prediction could, thus, be combined semantic part-segmentation and the robot can try perform the action on the segmented part (or area) where the affordance-guaranteeing fraction of pixels had been found.

A strength of our method is that it operates on 2D images and can be applied on all kinds of scenes, even in presence of light-absorbing, transparent and reflecting materials where structured light can not be used. Future work have to focus on improving the prediction of small structures, possibly by accounting for them in the cost function.

\printbibliography

\end{document}